# COMPREHENSIVE DATASET FOR URBAN STREETLIGHT ANALYSIS

*Eliza Femi Sherley S , Sanjay T, Shri Kaanth P, Jeffrey Samuel S*

Department of Information Technology, Madras Institute of Technology Campus,

Anna University, Chennai ,India

## ABSTRACT

This article includes a comprehensive collection of over 800 high-resolution streetlight images taken systematically from India's major streets, primarily in the Chennai region. The images were methodically collected following standardized methods to assure uniformity and quality. Each image has been labelled and grouped into directories based on binary class labels, which indicate whether each streetlight is functional or not. This organized dataset is intended to make it easier to train and evaluate deep neural networks, allowing for the creation of pre-trained models that have robust feature representations. Such models have several potential uses, such as improving smart city surveillance systems, automating street infrastructure monitoring, and increasing urban management efficiency. The availability of this dataset is intended to inspire future research and development in computer vision and smart city technologies, supporting innovation and practical solutions to urban infrastructure concerns. The dataset can be accessed at https://github.com/Team16Project/Street-Light-Dataset/.

## OBJECTIVE

The key objective of this dataset is to provide a diverse and comprehensive collection of streetlight images, specifically selected for computer vision-based inspection, monitoring, and maintenance of streetlights. This dataset, by giving a diverse group of images, aims to make the task easier to design and evaluate robust machine learning models and algorithms for streetlight identification, categorization, and operating status assessment.

## MOTIVATION AND SIGNIFICANCE

Effective streetlight management is essential in the rapidly evolving landscape of smart cities for improving urban infrastructure, increasing public safety, and optimizing energy consumption. Traditional streetlight monitoring systems frequently include manual inspections, which are time-consuming, labour-intensive, and prone to human mistake. The emergence of computer vision and machine learning technology provides a potential to transform this domain by automating inspection and maintenance procedures.

This dataset [1][3][7][8] holds significance for a wide range of reasons:

**Enhanced Urban Management**: This dataset facilitates the construction of automated monitoring systems, resulting in more efficient urban management techniques. Automated systems can immediately identify and report non-functional streetlights, resulting in rapid repairs with less downtime.

**Energy Optimization**: Efficient monitoring can assist in finding malfunctioning or energy-inefficient streetlights, resulting in improved energy usage and cost savings for local governments.

**Public Safety**: Well maintained streetlights are essential for public safety. Automated inspection systems can ensure that streetlights work properly, increasing the safety of metropolitan surroundings at night.

**Research and Development**: The dataset is an invaluable resource for computer vision, machine learning, and smart city researchers. It provides an exhausting set of images for creating and testing unique algorithms, stimulating creativity and setting limits in numerous fields.

**Accessible Resources**: The dataset is hosted on GitHub, making it easily accessible to the global research community, promoting collaboration and data exchange. It provides a uniform baseline for comparing the performance of various computer vision models and algorithms for urban infrastructure monitoring.

This dataset plays an essential role in enhancing urban planning and infrastructure management since it addresses the needs of smart city applications while also aiding the development of modern machine learning solutions.

## DATASET DESCRIPTION

The dataset discussed in this article includes about 800 high-resolution streetlight photos gathered in real time from urban regions in and around Chennai. Each image has a classification label that indicates whether the streetlight is operational ("working"), faulty ("flickering"), or inoperable ("not working"). The dataset aims to provide a wide and representative collection of streetlight settings observed in metropolitan scenarios, making it an important resource for training deep neural networks. These pre-trained models are suitable for smart city applications, including CCTV surveillance and street infrastructure monitoring.

## 3.1 COLLECTION METHODOLOGY

The images were taken manually with a camera at various sites within and outside of the Chennai region. The majority of the images were captured using a smartphone camera with an aperture of f/1.8 and a resolution of 12 MP (4000 x 3000 pixels). The data was collected at various times during the night to account for differences in ambient illumination and weather conditions, improving the dataset's durability and applicability to real-world circumstances.

## 3.2 DATASET FORMAT AND AVAILABILITY

The dataset is classified into three main categories based on the streetlights operational status: "working," "flickering," and "not working." It contains images and around 20 MP4 videos of flickering streetlights that give rich data for a variety of machine learning tasks. The images have been structured into a hierarchical folder system, with different directories for each class label, making them easier to use and navigate for researchers and practitioners. Individuals and organizations interested in downloading the dataset can do so using the GitHub repository https://github.com/Team16Project/Street-Light-Dataset. The images in Figure 1 are indicative of the collection.

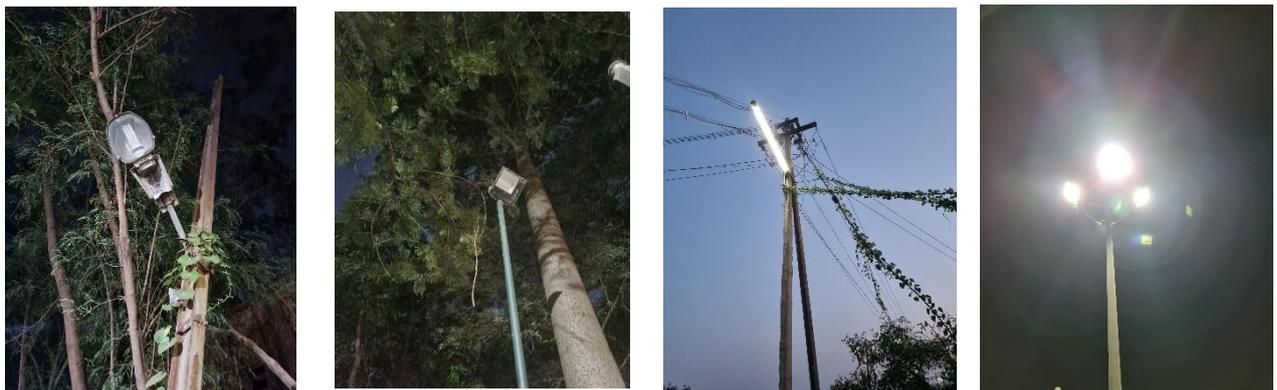

**Figure 1: Sample Images of the Dataset**

## 3.3 DATASET CHARACTERISTICS

The dataset provides a detailed classification of streetlight images and videos, organized by their operational status. The table below summarizes the class labels, the number of instances for each class, and their respective file formats.

**Table 1: Dataset Composition**

| Class Label | Count | File Format |
|---|---|---|
| Working | 533 images | .jpg |
| Not Working | 295 images | .jpg |
| Flickering | 21 videos | .mp4 |

## 4. EXPERIMENTAL BASE MODEL AND ARCHITECTURE

## 4.1 RESNET-18

In this study, we used ResNet-18 as the base model to train and validate our dataset of streetlight images[2][4][5][6]. ResNet-18 is a convolutional neural network (CNN) architecture known for its superior performance in image classification tasks[9]. It has 18 layers and uses residual learning, with shortcut connections that eliminate the vanishing gradient problem prevalent in deep networks. These shortcut connections allow the network to learn residual mappings, making it easier to train deeper models while maintaining performance. Figure 2 shows the base model architecture implemented for street light fault identification

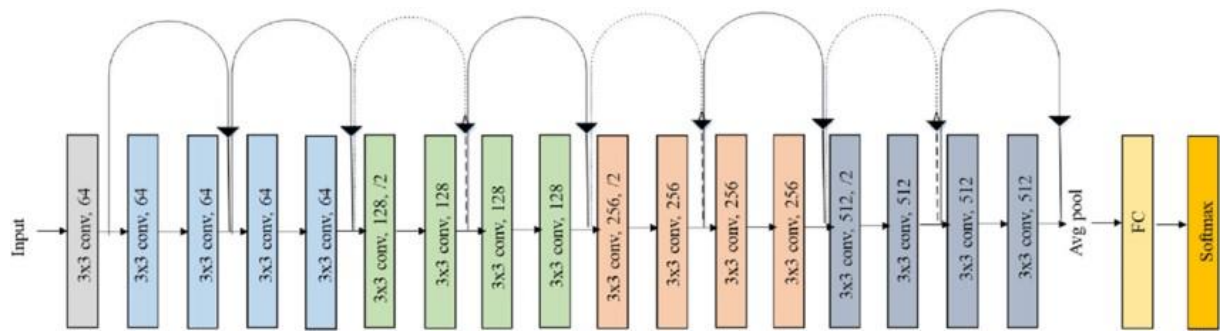

**Figure 2: RESNET -18 Architecture for Street Light Fault Identification**

ResNet-18 begins with basic operations such as convolution and pooling, progressively extracting increasingly complex features from the input images. Its lightweight design and robust performance make it a popular choice for a variety of computer vision tasks, including image classification, object detection, and semantic segmentation. By using ResNet-18, this study aims to develop a highly accurate model for classifying the operational status of streetlights, enhancing the utility of our dataset for smart city applications. By testing and training the dataset (Table 1) with ResNet-18, this study aims to develop pre-trained models that can be utilized in smart city systems for automated streetlight monitoring and maintenance.

## 4.2 RESULTS AND DISCUSSION

After testing and training the dataset (Table 1) with ResNet-18, this study achieved a training accuracy exceeding 86.19%. This demonstrates significant performance and effectiveness in accurately classifying images within the dataset using the model. Figure 3(a) represents the training and validation loss and Figure 3(b) represents the training and validation accuracy using RESNET-18 architecture.

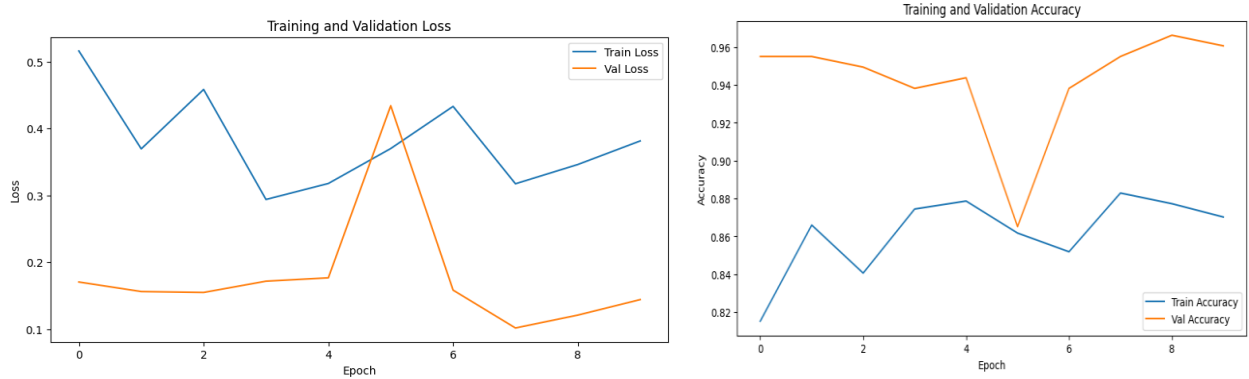

**Fig 3(a-b). Training/ Validation loss and accuracy using Resnet-18 architecture**

After training the model, the test images are classified as either working or not working. The sample prediction is shown in Figure 4

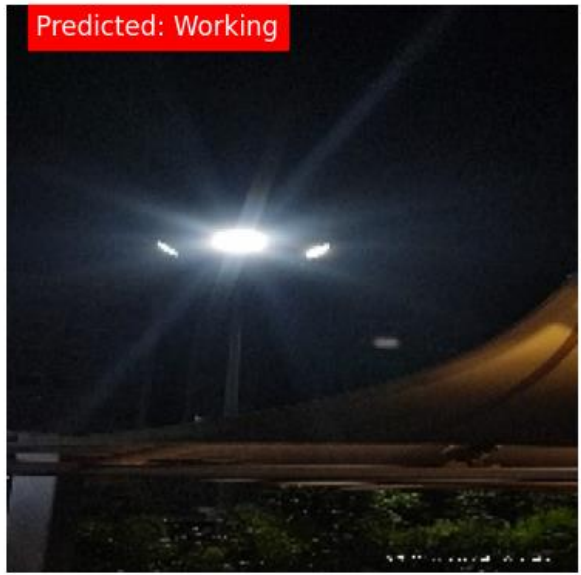 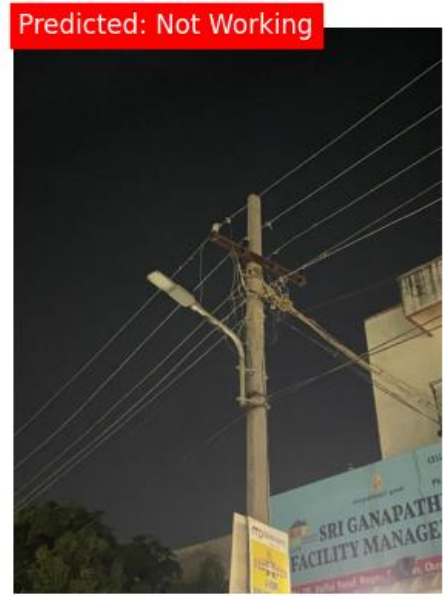

**Figure 4- Sample street light fault identification labelled as working/not working**

## 5. ACKNOWLEDGEMENT

We would like to extend our gratitude to the Greater Chennai Corporation (GCC) for allowing us to collect the dataset in public areas within their jurisdiction. Additionally, we are immensely thankful to the Department of Information Technology at Madras Institute of Technology Campus, Anna University for their guidance and support throughout this work. Their valuable insights have significantly contributed to the success of this work.